\documentclass[11pt]{article}

\usepackage[preprint]{acl}

\usepackage{times}
\usepackage{latexsym}

\usepackage[T1]{fontenc}

\usepackage[utf8]{inputenc}

\usepackage{microtype}

\usepackage{inconsolata}

\usepackage{graphicx}

\usepackage{microtype}
\usepackage{graphicx}
\usepackage{subcaption}
\usepackage{booktabs} 

\usepackage{amsmath}
\usepackage{amssymb}
\usepackage{mathtools}
\usepackage{amsthm}
\usepackage{xspace}
\usepackage{multirow}
\usepackage{xcolor}
\usepackage{soul}
\usepackage{pifont}

\usepackage[capitalize,noabbrev]{cleveref}
\crefname{section}{Section}{Secs.}
\Crefname{section}{Section}{Sections}
\Crefname{table}{Table}{Tables}
\crefname{table}{Tab.}{Tabs.}

\theoremstyle{plain}

\theoremstyle{definition}

\theoremstyle{remark}

\newcommand{\ours}{\textbf{VEN-VL}\xspace}
\definecolor{gray_tab}{RGB}{235, 235, 235}
\definecolor{blue_tab}{RGB}{207, 220, 251}
\definecolor{lblu_tab}{RGB}{225, 235, 246}
\definecolor{oran_tab}{RGB}{252, 242, 237}

\title{VEN-VL: A Visual Ensemble MoE Framework for \\ Effective and Efficient Multi-Modal Understanding}

\author{Yinghao Wu\footnotemark[1],
Zhuoyan Luo\footnotemark[1], \\
\textbf{Yiyao Yu}, 
\textbf{Zhaojian Yu},
\textbf{Yujiu Yang}\footnotemark[2], 
\textbf{Xiao-Ping Zhang}\footnotemark[2] \\
Tsinghua University \\
\texttt\small{\small yh-wu23@mails.tsinghua.edu.cn, \small luozy23@mail.tsinghua.edu.cn} \\ 
\small{Code:  \url{https://github.com/RobertLuo1/VEN-VL}}} 

\begin{document}
\maketitle

\renewcommand{\thefootnote}{\fnsymbol{footnote}}
\footnotetext[1]{Equal contribution.}
\footnotetext[2]{Corresponding author.}

\begin{abstract}

Despite the remarkable progress achieved by recent efficient methods in accelerating multimodal understanding, they still suffer from noticeable performance degradation.
Their emphasis on the high compression ratio of a single visual clue and reliance on the heuristic pruning strategy with coarse attention alignment incurs a bottleneck on the information \textit{capacity} and \textit{density} of visual tokens. 
Addressing this limitation, we propose VEN-VL, a visual ensemble MoE framework for effective and efficient perception following the \textit{enrich then compact} principle. Specifically, we first enrich the information capacity by unifying the visual representations of different perspectives, and then progressively compact it with adaptive routers in specialized visual experts to enhance the information density. Furthermore, we incorporate the reconstruction ability of vanilla structure via explicit visual supervision, facilitating crucial information preservation. 
Experimental results demonstrate our superiority in complex visual tasks with few information-condensed tokens, which effectively bridges the gap between performance and efficiency. 

\end{abstract}

\section{Introduction}
Recent success of Large Language Models (LLM)~\cite{openai2023gpt4, llama3, qwen3} has facilitated a significant advancement in Large Visual-Language Model (LVLM)~\cite{llava, instructblip, Qwen2vl, minigpt4}. 
The remarkable scalability of LVLMs, especially with Mixture of Experts (MoE) architectures~\cite{llava-moe, deepseek-vl2, kimivl, molmo, gemini25}, demonstrates strong potential for complex visual understanding in real-world interactions. 
Due to the substantial computational overhead of LVLM deployment, efficiency and effectiveness are key considerations in model design.

\begin{figure}[t]
  \begin{center}
    \centerline{\includegraphics[width=\columnwidth]{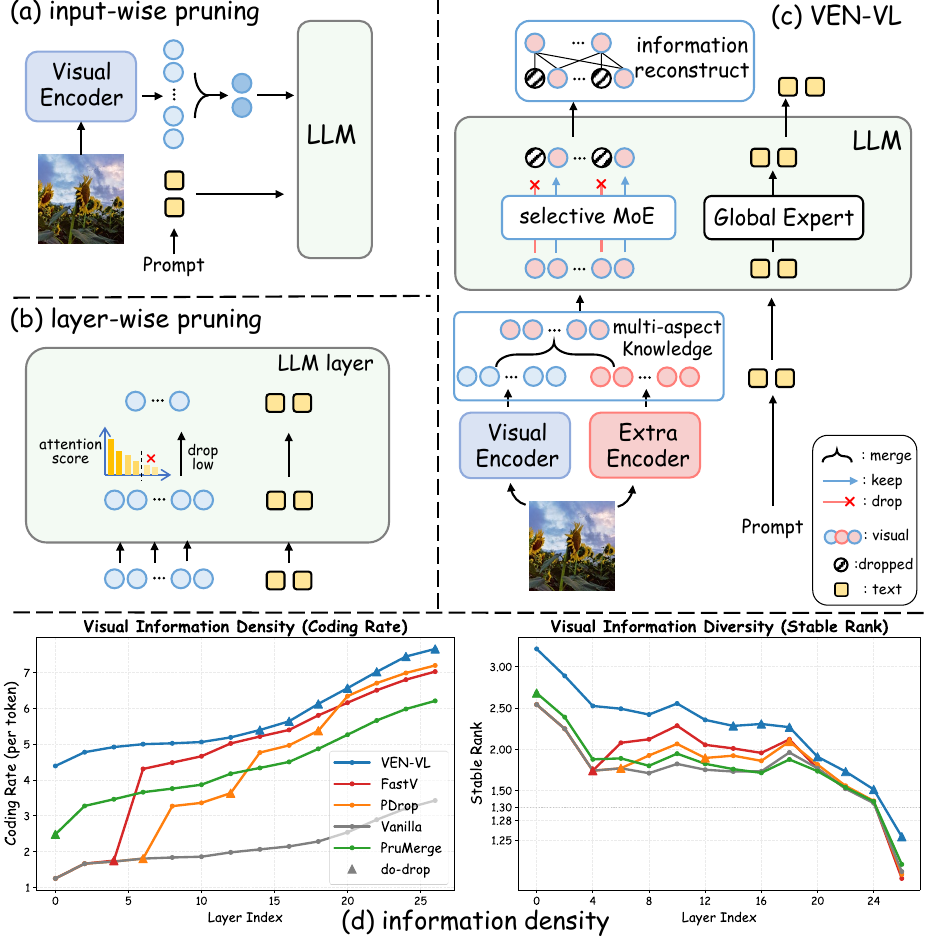}}
    \caption{
      \textbf{Comparison with different paradigms.} Existing methods typically rely on either input-wise compression of single-aspect features (a) or heuristic layer-wise pruning based on coarse attention alignment (b), both of which suffer from the limitation of information capacity and density. In contrast, we propose \ours (c), which first unifies multi-perspective visual representations via MKE to boost capacity, and then performs progressive refinement with dynamic routers of HTE in the LLM. As visualized in (d), \ours demonstrates superior information density of each layer compared to previous efficient methods.
    }
    \vspace{-35pt}
    \label{fig:teaser}
  \end{center}
\end{figure}


Considering the quartic complexity is highly associated with the token sequences, a straightforward solution is visual token pruning. 
Following this, a variety of approaches are presented to explore effective mechanisms for visual redundancy reduction while maintaining valuable tokens. It can be generally divided into two aspects in terms of their design principles. 
On the one hand, the \cref{fig:teaser}(a) depicts the methods~\cite{blip2, voco, llava-mini, prumerge, visionzip} that compress the single vision features into a compact representation before feeding them into the LLM. 
On the other hand, as shown in \cref{fig:teaser}(b), the counterparts~\cite{pyramiddrop, fastv, sparsevlm, star} prune visual tokens at the layer of LLM based on the attention map.

Despite these progresses in model acceleration, they both suffer from the inevitable performance degradation. 
We speculate it stems from two key limitations. 
In terms of the \textit{input-wise pruning}, the emphasis of the exaggerated compression rate on the single-stream vision features fails to establish the completed semantic modeling. This decreases the information capacity and diversity significantly (as indicated in \cref{fig:teaser}(d)), thereby becoming a bottleneck for subsequent multimodal understanding in LVLMs. 
For \textit{Layer-wise pruning}, we identify that their information density for complex visual understanding still falls below the anticipation. It suggests that attention-map-based selection mistakenly drops the tokens with condensed information. It is associated with the naive information accumulation on coarse visual-text alignment, which potentially incurs shifting emphasis beyond the valuable visual tokens, \textit{i.e.}, fine-grained semantics required for later layers to be fully comprehended.
Hence, a central question arises: \textbf{\textit{Can we enable LVLMs to realize superior performance with fewer but highly information-condensed visual tokens?}}

To explore this and break through the limitations above, the \textit{capacity} and \textit{density} of visual information are supposed to be jointly considered. 
Intuitively, the former determines the upper bound of the model's multimodal understanding performance, whereas the latter plays a pivotal role in the precise and efficient visual perception. 
Therefore, we conduct a pilot study to quantify them via stable rank~\cite{stablerank} and coding rate~\cite{coderate} in \cref{fig:teaser}(d) respectively, and present \ours, a \textbf{V}isual \textbf{EN}semble MoE framework for effective and efficient \textbf{V}isual-\textbf{L}anguage understanding.
To achieve improvement both on the information capacity and density, we formulate the design of \ours following the \textit{enrich then compact} principle.
Firstly, as described in \cref{fig:teaser}(c), instead of simply compressing on the single stream visual feature, the Multi-Aspect Knowledge Ensemble (MKE) is leveraged for \textit{enriching} semantic representations by different perspectives of visual clues with structure-aware compression. 
It not only enhances the comprehensive representations of the given visual cue but also suppresses its redundancy, where the \textit{capacity} and diversity of valuable visual information can be largely boosted. 
Secondly, as for the information density, we propose Hierarchical Token Ensemble (HTE), which seamlessly integrates the inherent specificity of MoE to \textit{compact} the visual distribution progressively. 
With the design of the adaptive router to dynamically decouple intricate semantics into different experts for fine-grained processing and selection, \ours demonstrates superiority in information \textit{density}, as presented in \cref{fig:teaser}(d).
Moreover, to further provide indicative guidance on the token selection process, we introduce the Structure Information Preservation (SIP). 
SIP embodies the model with reconstruction ability on the pruned token via semantic propagation then supervision, thus ensuring the retained tokens restore vanilla condensed information to a great extent.

We conduct extensive experiments on prevalent multi-modal benchmarks, \textit{e.g.}, MMBench~\cite{mmbench}, TextVQA~\cite{textvqa} and MME~\cite{mme}, etc. 
Experimental results demonstrate that \ours excels at complex visual understanding tasks compared to previous efficient and MoE-based LVLMs with merely \textbf{7.5$\%$} tokens remained, which effectively bridges the gap between performance and efficiency. Moreover, we further validate the strong compatibility of our design with the current MoE framework~\cite{kimivl}, demonstrating its considerable potential for scaling to larger models. Overall, our contributions can be summarized as follows:
\begin{itemize}
    \item Based on the observation that the information \textit{capacity} and \textit{density} play an essential role in bridging the gap between performance and efficiency, we propose the \textit{enrich then compact} principle that guides the design of the visual ensemble MoE framework for visual-language understanding. \ours first formulates an enriched and compact visual space (MKE), and then progressively refines the visual distribution by adaptive routers (HTE).  
    \item We introduce the Structure Information Preservation (SIP) to provide guidance on the token selection process via assigning the reconstruction ability of the original visual cues.
    \item \ours achieves superior performance in complex multi-modal understanding tasks, \textit{e.g.}, $+8.1\%$ in TextVQA and $+5.2\%$ in MMB with fewer but more information-condensed visual tokens ($\sim 10\%$).
\end{itemize}

\section{Related Work}

\paragraph{Efficient LVLM.}
Efficiency is one of the key topics in LVLM, where the design can be divided into two main categories. 
1) \textit{Input visual representations compression}: early methods~\cite{instructblip, du2022glm, minigpt4, blip2} leverage Q-Former~\cite{blip2} to compress tokens into fixed-length 1D learnable queries. VoCo-LLaMA~\cite{voco} utilizes LLM for distillation, which converts the visual features into one specific token. LLava-Mini~\cite{llava-mini} introduces the modality pre-fusion module for visual-text interaction, achieving minimal tokens fed into the LLM backbone. LLaVA-PruMerge~\cite{prumerge} and VisionZip~\cite{visionzip} directly perform redundancy reduction on the input visual features. Obviously, they overemphasize the exaggerated compression ratio, which faces the challenges in modeling the valuable spatial properties of the original visual representations and limited semantic information. 2) \textit{Inner visual token reduction}: These methods~\cite{fastv, pyramiddrop, atpllava, star} typically perform similar token pruning mechanisms relying on the attention map of each layer in LVLM. However, we observe that the pre-defined threshold-based token selection without guidance is coarse, which exhibits vulnerability in handling diverse visual scenarios, leading to inevitable information density loss.

\paragraph{Mixture of Experts in LVLM.}
To unleash the scaling potential of LVLM and simultaneously avoid extensive computational overhead, Mixture of Experts (MoE)~\cite{shazeer2017outrageously, eigen2013learning} is adopted to enable routers for specific parameter activations. Prior work LLaVA-MoE~\cite{llava-moe} advances LLava~\cite{llava} with the proposed MoE-Tuning strategy, empowering the visual-text understanding capacity. Moreover, several approaches~\cite{kimivl, deepseek-vl2, molmo} demonstrate a crucial step forward in MoE-based LVLM training. Although the MoE largely boosts the performance with sparse activated parameters, it still incurs huge memory consumption considering the quartic costs of the visual tokens. Therefore, we endeavor to excavate the potential of fewer but information-dense visual tokens by incorporating the natural selective properties of MoE, which leaps towards bridging the performance and efficiency gap.

\section{Method}
\begin{figure*}[ht]
  \begin{center}
    \centerline{\includegraphics[width=0.8\linewidth]{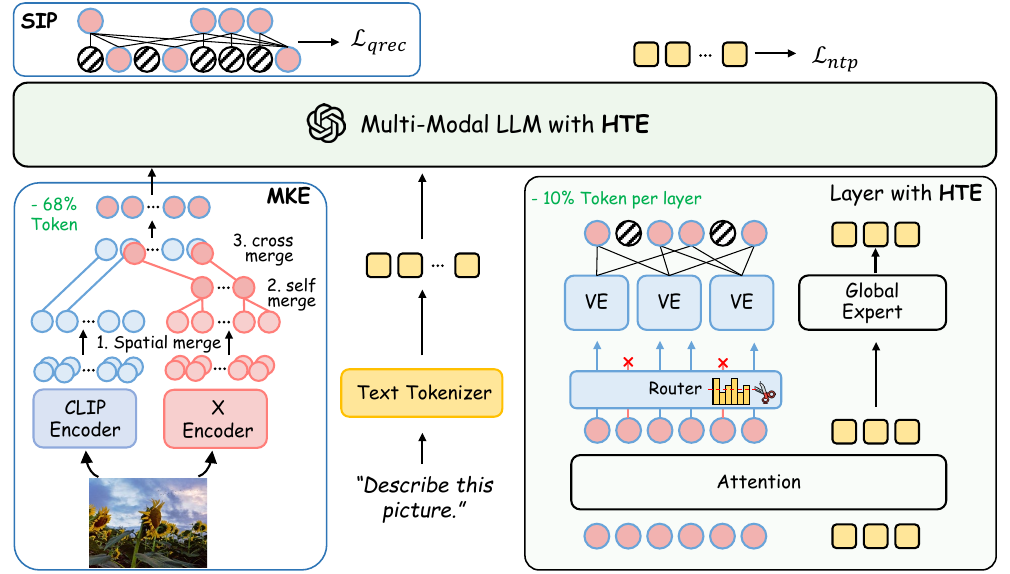}}
    \caption{
      \textbf{Overview of \ours.} The model incorporates three components: MKE, HTE, and SIP. The MKE first extracts visual features of different perspectives and ensembles them into a compact and unified representation through spatial- and cross-merging, which enhances semantic diversity and reduces redundancy. Subsequently, the HTE benefits from the inherent specialty of MoE to perform fine-grained token selection with dynamic routers. It progressively refines the visual distribution across layers, gradually increasing the information density. Finally, SIP is used to provide inductive guidance on the reduction process, which empowers the model with reconstruction ability of the structural information.
    }
    \vspace{-20pt}
    \label{fig:framework}
  \end{center}
\end{figure*}

We introduce \ours, a MoE framework with comprehensive visual ensemble.  
As depicted in \cref{fig:framework}, it is composed of three main components. Following the \textit{enrich then compact} principle, we first present Multi-Aspect Knowledge Ensemble, termed as MKE in \cref{sec:mske}, which enrich semantic diversity while alleviating the redundancy of the visual representation. Subsequently, based on the knowledge-enhanced input, the hierarchical visual token ensemble strategy in \cref{sec:hvte} incorporates the adaptive router in MoE to improve the information density of visual tokens gradually. Finally, in \cref{sec:saie}, we elaborate on the proposed structural information preservation.

\subsection{Architecture}
\label{sec:arch}
The overall architecture of \ours is presented in \cref{fig:framework}. Given an image $\mathcal{I} \in \mathbb{R}^{H \times W \times 3}$, where $H$ and $W$ denote the original height and width. The Multi-Aspect Knowledge Ensemble (MKE) with a dual-branch visual encoder is to extract different perspectives of vision features, \textit{i.e.}, semantic-centric $\mathcal{F}_{main} \in \mathbb{R}^{H_{1} \times W_{1} \times D_1}$ and fine-grained details-amplified $\mathcal{F}_{extra} \in \mathbb{R}^{H_{2} \times W_{2} \times D_2}$. To establish a unified visual space while suppressing the redundancy, MKE then performs merging mechanisms to generate a joint embedding $\mathcal{V} \in \mathbb{R}^{L_{v} \times D}$, where $D$ represents the hidden size of LLM.
Subsequently, Hierarchical Token Ensemble (HTE) is applied to take $\mathcal{V}$ as the input and increase the information density of visual tokens progressively via incorporating adaptive routers of MoE. By hierarchical compacting the visual tokens, it ultimately derives pruned but semantic-condensed tokens $\mathcal{V}^{c} \in \mathbb{R}^{L^{c}_{v} \times D}$ where $L^{c}_{v}$ $\ll$ $L_{v}$, for multi-modal understanding. Further, to consolidate the ensemble process of the visual tokens, Structural Information Preservation (SIP) is proposed to reconstruct the original structure of the input feature $\mathcal{V}$ by semantic propagation with visual supervision.

\subsection{Multi-Aspect Knowledge Ensemble}
\label{sec:mske}
A family of existing efficient methods~\cite{prumerge, llava-mini, voco} focuses on the large compression ratio of the visual representation from a single stream, \textit{e.g.}, CLIP. 
Despite the acceleration of the whole pipeline, it struggles with the complex multi-modal understanding task, especially the one that necessitates diverse interpretations of visual cues. The rationale behind falls on the significant degradation of the information capacity and diversity. 
Therefore, we propose the Multi-Aspect Knowledge Ensemble (MKE) with two key incentives to tackle the issues. 
First, considering that CLIP exhibits insufficient fine-grained details modeling due to the optimization on image-level visual-linguist alignment, we incorporate an additional visual encoder with heterogeneous representation to enable comprehensive semantics activation. 
Moreover, we observe that the aforementioned approaches employ a new set of learnable embeddings for information integration, disregarding the valuable structural information. Instead, as illustrated in \cref{fig:framework}, we strive to maintain the vanilla visual relationship and suppress the redundancy with structural aware compression. With sequential merging on the visual representations of different perspectives, it well establishes the compact and unified visual space.

Specifically, given an input image $\mathcal{I}$, we leverage CLIP to derive the semantic-aligned features $\mathcal{F}_{main} \in \mathbb{R}^{N \times D}$. Simultaneously, we enrich the information capacity with the representation $\mathcal{F}_{extra} \in \mathbb{R}^{N \times D}$ at a fine-grained level, which is extracted from the auxiliary encoder. 
Although the dual-stream features primarily emphasize different aspects, the direct concatenation of them fails to formulate a complete holistic semantics due to the substantial noise accumulation and huge computational overhead. 
In order to ensemble the crucial information seamlessly with efficiency and avoid the overwhelming disruption by the redundancy, we adopt a structure-aware compression strategy with sequential merging.
Firstly, we compress on the spatial axis of both visual representations, which aims at transforming the locality into the channel dimension. It initially alleviates the redundancy in the \textit{local-level} while preserving the original 2D grid topology, which can be formulated as:
\begin{equation}
\hat{\mathcal{F}} = \Psi(\mathcal{F}, r) \in \mathbb{R}^{\frac{H}{r} \times \frac{W}{r} \times (C \cdot r^2)},
\end{equation}
where $\Psi$ denotes the transformation and $r$ is the downsampling ratio. Then, as we intend to retain the critical structure information and facilitate the holistic understanding of the visual cues, we keep $\mathcal{F}_{main}$ unchanged and employ self/cross-merge on the $\mathcal{F}_{extra}$ for redundancy suppression and semantic complementary in the \textit{global-level}. More detailedly, inspired by~\cite{tome, avit, spvit}, we introduce a bipartite merge operator $\Phi(\mathbf{X}, \mathbf{Y}, k)$. It divides the input into a source set $\mathbf{Y}$ and a target set $\mathbf{X}$, then identifies the $k$ edges with the top $k$ cosine similarity $\cos(\mathbf{x}_i, \mathbf{y}_j)$, and aggregates the matched pairs via weighted averaging:
\begin{equation}
\Phi(\mathbf{X}, \mathbf{Y}, k) = \alpha \mathbf{x}_i + (1-\alpha) \mathbf{y}_j
\end{equation}
With the pre-defined operator, we decouple $\hat{\mathcal{F}}_{extra}$ into two disjoint features $\hat{\mathcal{F}}^A_{extra}$ and $\hat{\mathcal{F}}^B_{extra}$, and obtain the more compact representation:
\begin{equation}
\tilde{\mathcal{F}}_{extra} = \Phi(\hat{\mathcal{F}}^A_{extra}, \hat{\mathcal{F}}^B_{extra}, k_{self})
\end{equation}
Finally, it is further projected into $\mathcal{F}_{main}$ via cross-merge for semantic complementary, which amplifies the original information with fine-grained visual cues:
\begin{equation}
\mathcal{V} = \Phi(\hat{\mathcal{F}}_{main}, \tilde{\mathcal{F}}_{extra}, k_{cross})
\end{equation}
In this manner, we formulate the unified and compact visual space $\mathcal{V}$ by the semantic ensemble with structure-aware compression, significantly enriching the information capacity and diversity.

\subsection{Hierarchical Token Ensemble}
\label{sec:hvte}
While MKE establishes a semantic-\textbf{\textit{enriched}} visual space at the input level, \textbf{\textit{compacting}} these visual tokens efficiently and effectively within the deep layers of the LLM afterwards remains critical. Existing layer-wise pruning methods typically rely on attention maps for token selection. However, this approach may mistakenly drop tokens with condensed information due to the naive visual-text alignment in the early layer, \textit{e.g.}, those corresponding to small objects or subtle texture representations.
To address the issues, we propose the Hierarchical Token Ensemble (HTE), which benefits from the inherent specificity of the MoE architecture to decouple intricate semantics and progressively refine the visual distribution by the selective mechanism. 
Specifically, as shown in \cref{fig:framework}, we first adopt a specialized visual MoE design to process the fine-grained semantics separately without deteriorating the general reasoning capabilities of LLM. Then, we retain the original pre-trained FFN as the Global Expert ($E_{global}$), which operates on both visual and textual tokens to maintain general semantic knowledge.
Simultaneously, we introduce a bank of lightweight experts $\mathcal{E} = \{E_k\}_{k=1}^N$, which are activated exclusively for visual tokens to capture specific visual semantics. 
Supposing that $\mathbf{x} \in \mathbb{R}^D$ denotes the input hidden state. The output $\mathbf{y}$ integrates the dense global processing with sparse visual expertise:
\begin{equation}
\mathbf{y} = \mathbf{x} + E_{global}(\mathbf{x}) + \mathbb{I}_{vis}(\mathbf{x}) \cdot \sum_{i \in \mathcal{K}} w_i(\mathbf{x}) E_i(\mathbf{x})
\end{equation}
where $\mathbb{I}_{vis}$ is the modality indicator, and $\mathcal{K}$ signifies the set of selected indices corresponding to the experts. The routing weights $w(\mathbf{x})$ and indices $\mathcal{K}$ are derived via a Top-$k$ gating mechanism:
\begin{equation}
w(\mathbf{x}) = \text{Softmax}(\mathbf{x} W_r), \quad \mathcal{K} = \operatorname{TopK}(w(\mathbf{x})),
\end{equation}
where $W_r \in \mathbb{R}^{D \times N}$ represents the router projection. It dynamically guides visual tokens to the experts that are best suited for the particular semantic content.

It is worth noting that the utilization of the router is not only for computing the distribution of routers, but also serves as a judge for token saliency. Specifically, we observe that the routing activation reflects the distinctiveness of the visual information within a token. 
On one hand, tokens that exhibit a high saliency score, \textit{i.e.}, $w_i(\mathbf{x})$, for specific experts necessitate oriented processing to fully interpret the intricate, fine-grained semantics. In contrast, that is typically neglected by the previous methods that rely on the coarse attention map for token pruning. 
On the other hand, tokens with low saliency scores are likely to contain trivial or redundant information that has already been captured by earlier layers.

Therefore, A reduction-gated mechanism to refine the visual distribution is leveraged. We define the importance score $s(\mathbf{v})$ of a visual token $\mathbf{v}$ as its maximum routing activation:
\begin{equation}
    s(\mathbf{v}) = \max_{k \in \mathcal{K}} g_k(\mathbf{v})
\label{eq:token_score}
\end{equation}
At specified layers, we apply a threshold $\tau$ for pruning. The set of retained visual tokens $\mathcal{V}_{next}$ for the subsequent layer is denoted as:
\begin{equation}
\mathcal{V}_{next} = \{ \mathbf{v} \in \mathcal{V}_{curr} \mid s(\mathbf{v}) > \tau \}
\end{equation}
By filtering tokens dynamically with their saliency instead of coarse attention alignment, HTE progressively increases the information density of the visual sequence across layers. This ensures that the deep-layer computation is concentrated solely on the highly information-condensed tokens required for precise multi-modal understanding.

\subsection{Structure Information Preservation}
\label{sec:saie}
As the visual modality contains a wide spectrum of features including spatial relationships, textures, and complex compositions, the visual token selection strategy should maximize the preservation of information capacity and density.
However, we notice that existing efficient methods~\cite{pyramiddrop, visionzip} tend to perform layer-wise token pruning in a heuristic manner. That is simply relying on the attention map as the only criteria for reduction. Yet, it exhibits vulnerability in information maintenance and even exacerbates the loss due to the lack of explicit visual supervision.

Therefore, we propose Structure Information Preservation (SIP) to provide inductive guidance on the token selection process. It endows the model with reconstruction ability of the original structure $\mathcal{V}$ via semantic propagation, which ensures the remaining visual tokens largely preserve vanilla condensed information. 
Inspired by~\cite{uvcom, weaklysupervised, random_walk}, we adopt a parameter-free approach to avoid the model developing excessive reliance on the learnable parameters \textit{e.g.}, cross attention or self-attention mechanism, for reconstruction adaptation instead of concentrating on the information-condensed visual tokens. 
Specifically, we first re-assign the pruned token embedding $\mathcal{V}^{p}$, with RoPE for structural property initialization. Then, the remaining information-condensed tokens $\mathcal{V}^{c}$ iteratively propagate the holistic semantics to $\mathcal{V}^{p}$ in a shared space until convergence. Formally, at $0$-th iteration, the affinity $\mathcal{Z}$ is formulated by scaled dot product similarity: $\mathcal{Z}=\lambda \mathcal{V}^{c}_{(0)}(\mathcal{V}^{p}_{(0)})^{\top}$, where $\lambda$ denotes the scaling factor. To mitigate the noise accumulation during interaction and prevent causing redundancy, the top-$k$ activations with high saliency scores are selected:
\begin{equation}
\tilde{\mathcal{Z}} =
\begin{cases}
\mathcal{Z}, & i \in \operatorname{TopK}\!\left(\mathcal{Z}\right), \\
0, & \text{otherwise},
\end{cases}
\end{equation}
While at $t$ iterations, $\mathcal{V}^{c}_{(t)}$ is updated jointly by the original representation $\mathcal{V}_{(0)}^{t}$ and structure-aware token embedding $\mathcal{V}_{(t-1)}^p$ from previous iteration:
\begin{equation}
\mathcal{V}^{c}_{(t)} = \omega \, \mathrm{Norm}(\tilde{\mathcal{Z}})^{\top} \mathcal{V}^{p}_{(t-1)} + (1 - \omega)\, \mathcal{V}^c_{(0)} ,
\end{equation}
Subsequently, it is projected into $\mathcal{V}^{p}_{(t)}$:
\begin{equation}
\mathcal{V}_{(t)}^{p} = \omega \tilde{\mathcal{Z}}\mathcal{V}^{c}_{(t)} + (1 - \omega)\mathcal{V}^{p}_{(0)} ,
\end{equation}
where $\omega$ signifies the level of semantic propagation. After $t$ iterations, we obtain the reconstructed visual representation $\mathcal{V}^{p}$ that retains the original structural information. To further provide guidance on refining the visual distribution \cref{sec:hvte} for precise information preservation, we incorporate visual supervision between the $\mathcal{V}^{p}$ and $\mathcal{V}$ via fine-grained alignment:
\begin{equation}
\mathcal{L}_{qrec} = \sum||\left(\operatorname{Quant}(\mathcal{V}^{p}) - \operatorname{Quant}(\mathcal{V})\right)||_{2},
\end{equation}
where $\operatorname{Quant}\left(\right)$ indicates quantization~\cite{fsq}, which is utilized to prevent overwhelmingly regressing the exact value~\cite{ross}. With guidance, the model is encouraged to accumulate structural information more effectively and to precisely alleviate visual redundancy, thereby improving both information capacity and density.

\section{Experiment}

\subsection{Experiment Setup}

Detailed descriptions of the implementation, training strategies, along with the datasets and benchmarks, are provided in the Appendix \cref{subsec:exp_setup}.

\begin{table*}[ht]
\centering
\caption{
\textbf{Benchmarked results with SoTA MLLMs}. Compared with previous efficient LVLM methods, \textit{e.g.}, token-pruning based and lightweight design, our \ours~demonstrates superiority in versatile visual understanding tasks with fewer but information-condensed tokens across different scales of LLM backbone. Optimal and sub-optimal performance is signified by \textbf{Bold} and \underline{underline}. Note that, 
 $^\dag$: denotes the reproduced results using the official code. P/Sum indicates perception scores and the total of perception and cognition metrics.}
\vspace{-10pt}
\renewcommand{\arraystretch}{0.9}
\resizebox{1.\linewidth}{!}{
\begin{tabular}{ccccccc|cccccc}
\toprule[0.17em]
\multirow{2}{*}{\textbf{Method}} & \multirow{2}{*}{LLM} & \multirow{2}{*}{Retain Ratio ($\%$)} & \multicolumn{4}{c}{Image Question Answering} & \multicolumn{6}{c}{Benchmarks}     \\\cline{4-13}
                                 &                      &                        & GQA & SciQA  & TextVQA  & AI2D & SEED$^{\text{I}}$ & MME(P/Sum)  & MMB  & $\text{MMB}^{\text{CN}}$ & POPE & MMMU \\
\hline
LLaVA-1.5 & Vicuna-7B & 100$\%$  & \underline{62.0}  & 66.8 & 58.2 & 56.4 & 66.1 & 1510/1862 & 64.3 & 58.3   & 85.9 & 34.4 \\
Qwen-VL-Chat & Qwen-7B & $100\%$ & 57.5  & 68.2   & 61.5 & 57.7 & 65.4 & 1488/1848 & 60.6 & 56.7 & $-$  & 35.9 \\
MiniCPM-V2 & MiniCPM-2.4B & $100 \%$ & $-$  & 80.4 & \underline{74.3} & 64.8 & $-$ & $-$/1788 & 69.4 & 65.9 & 86.6 & 39.6 \\
InternVL2-2B & InternLM2-1.8B & $100 \%$ & $-$  & \underline{94.1} & 73.5& 74.4 & \underline{71.6} & $-$/1863 & 73.4 & \underline{70.9} & 85.3 & 34.9 \\
Qwen2-VL-2B & Qwen2-1.5B & $100 \%$ & $-$  & 78.1 & \textbf{79.9} & 74.1 & 71.5 & $-$/1890 & \underline{74.9} & $-$ & 78.8 & \textbf{41.7} \\
\midrule
\multicolumn{13}{c}{\textit{Large Visual-Language Model With Token Pruning}} \\
\midrule
PruMerge & Vicuna-7B & $\sim 5\%$   & $-$ & 68.5 & 56.0 & $-$ & $-$ & 1350/$-$ & 60.9 & $-$ & 76.3 & $-$ \\
PruMerge+ & Vicuna-7B & $\sim 25\%$ & $-$ & 68.3 & 57.1 & $-$ & $-$ & 1462/$-$ & 64.9 & $-$ & 84.0 & $-$ \\
LLaVA-Mini & Vicuna-7B & $\sim 10\%$ & 61.8 & 69.7 & 59.1 & $-$ & 60.2 & 1477/$-$ & 67.5 & $-$ & 85.3 & $-$ \\
FastV & Vicuna-7B & $\sim10\%$ & 46.1 & 51.1 & 47.8 & $-$ & 51.9 & $-$/1256 & 48.0 & $-$ & 48.0 & 34.0 \\
SparseVLM & Vicuna-7B & $\sim 10\%$ & 52.7 & 62.2 &51.8 & $-$  & 51.1& $-$/1589 & 56.2 & $-$ & 75.1 & 32.7 \\
PyramidDrop & Vicuna-7B & $\sim 10\%$ & 47.5 & 69.0 & 50.6 & $-$  & 64.3 & $-$/1561 & 58.8 & 58.5 & 86.0 & $-$ \\
VisionZip & Vicuna-7B & $\sim 10\%$ & 55.1 & 69.0 & 55.5 & $-$ &  53.4 & $-$/1676 & 60.1 & $-$ & 77.0 & 36.2 \\
\midrule
\multicolumn{13}{c}{\textit{Large Visual-Language Model with Lightweight Design}} \\
\midrule
$\text{TinyLLaVA}^\dag$ & Qwen1.5-0.5B& $100\%$ & 57.4  & 60.9   & 47.4 & $-$ & $-$ & 1196/$-$ & 55.0 & 52.4 & 83.7 & 31.6 \\
$\text{TinyLLaVA}^\dag$ & Qwen2.5-0.5B & $100\%$ & 58.3  & 59.1   & 49.2  & $-$ & $-$ & 1230/$-$ & 58.9 & 54.2   & 86.1 & 33.6 \\
$\text{TinyLLaVA}^\dag$ & Qwen1.5-1.8B & $100\%$ & 55.5  & 65.3   & 47.7 & $-$ & $-$ &1224/$-$ & 57.1 & 55.5   & 83.4 & 34.1 \\
Bunny-2B & Qwen1.5-1.8B & $100\%$ & 59.6  & 64.6   & 53.2 & $-$ & $-$  & 1300/$-$ & 59.1 & 58.5   & 85.8 & $-$ \\
Imp-2B & Qwen1.5-1.8B & $100\%$ & 61.9  & 66.1   & 54.5 & $-$ & $-$ & 1304/$-$ & 63.8 & 61.3   & 86.7 & $-$ \\
Mini-Gemini-2B  & Gemma-2B & $100\%$ & 60.7  & 63.1   & 56.2 & $-$ & $-$ & 1341/1653 & 59.8 & 51.3   & 85.6 & 31.7 \\
MoE-LLaVA-2B & Qwen-1.5-1.8B & $100\%$ & 61.5   & 63.1   & 48.0 & $-$ & $-$   & 1292/$-$ & 59.7 & 57.3   & 87.0 & $-$ \\
LLaVA-MOD & Qwen1.5-1.8B         & $100\%$ & 58.7   & 68.0   & 58.5 & $-$ & $-$  & 1352/$-$ & 66.3 & 61.9   & 87.0 & $-$ \\
LLaVA-MOD & Qwen2-1.5B & $100\%$ & 58.8  & 69.2 & 59.9 & $-$ & $-$ & 1384/$-$ & 68.9 & 64.4& \underline{87.2} & $-$ \\
$\text{TinyLLaVA}^\dag$ & Qwen2.5-1.5B & $100\%$  & \textbf{62.0}  & 72.0 & 57.4 & $-$ & $-$ & 1450/$-$ & 68.6 & 63.0   & 85.5 & 37.0 \\
\hline
\ours & Qwen3-0.6B & $\sim 7.5\%$ & 58.8  & 84.2 & 61.0 & 68.3 & 69.0 & 1320/1614 & 66.5 & 64.5 & \textbf{87.5} & 35.6 \\ 
\ours & Qwen2.5-1.5B & $\sim 10\%$& 60.4  & 90.5  & 65.5 & \underline{74.6} & 70.3& 1401/1736 & 74.1 & 69.8  & 85.5 & 37.9 \\ 
\ours  & Qwen3-1.7B & $\sim 7.5\%$ & 60.6  & \textbf{94.5}   & 73.7 & \textbf{76.9} & \textbf{72.5} & \textbf{1504}/\textbf{1984} & \textbf{76.6} & \textbf{76.5} & 86.0 & \underline{39.7} \\
\bottomrule[0.12em]
\end{tabular}
}
\vspace{-0.9em}
\label{table:main_results}
\end{table*}

\begin{table*}[t]
  \caption{\textbf{Comparison with previous token pruning methods across LVLMs.} Comprehensive evaluation with previous token pruning based LVLMs under fair settings, \textit{i.e.}, all adopt Qwen3-0.6B as the LLM backbone with the identical training data. Benefiting from the enriched and dense visual representation, \ours achieves superior performance with significantly fewer tokens. Vanilla refers to the standard LLaVA~\cite{llava} implementation. Tokens denote the ratio of retained visual tokens. }
    \vskip -0.1in
  \label{tab:exp2}
  \centering
  \resizebox{\textwidth}{!}{%
    \begin{tabular}{c|cc|cccccccc}
        \toprule
        Method & Token$\downarrow$ & FLOPs(G)$\downarrow$ &  MMMU&MME & MMB & AI2D & TextVQA & GQA  & SEED$^{\text{I}}$ & Avg.$\uparrow$ \\ 
        \midrule
        Vanilla & 100\%& 2330.9 &  32.1 &1318& 53.4 & 48.8 & \textbf{44.0} & \textbf{55.6}  &62.7 & 49.1 
\\ 
\hline
        PruMerge & 12.5\%& 226.9 &  28.1 &1292& 41.2 & 40.5 & 33.5 & 47.9  &49.1 & 40.9 
\\ 
        PruMerge+ & 20.0\%& 350.1 &  27.8 &1302& 43.2 & 42.3 & 34.0 & 47.6  &50.7 & 41.7 
\\ 
        FastV & 12.5\%& 527.4 &  25.3 &1276& 45.5 & 43.7 & 35.8 & 42.8  &50.3 & 41.3 
\\ 
        PyramidDrop & 12.5\%& 953.2 &  29.6 &1260& 46.2 & 45.0 & 37.9 & 45.0  &52.2 & 43.0 
\\ 
\hline
        \ours & 7.5\%& 558.0 &  \textbf{34.3} & \textbf{1398} & \textbf{55.8} & \textbf{50.3} & 42.1 & 54.8  & \textbf{63.5} & \textbf{50.1} 
\\ 
        \bottomrule
    \end{tabular}
  }
  \vskip -0.1in
\vspace{-5pt}
\end{table*}

\subsection{Main Result}

\paragraph{Comparison with the SoTA LVLMs.}
To validate the effectiveness of \ours, we conduct a comprehensive evaluation across diverse benchmarks. As shown in \cref{table:main_results}, \ours consistently outperforms state-of-the-art efficient and lightweight models with significantly fewer visual tokens. Specifically, compared to the strong baseline TinyLLaVA (implemented with Qwen2.5-1.5B), \ours achieves substantial gains of +8.1\% in TextVQA, +5.2\% in MMB, and +6.8\% in MMB$^{CN}$, despite retaining only $\sim$10\% of the visual tokens. This highlights a significant leap towards bridging the gap between performance and efficiency. Moreover, the results across varying LLM backbones demonstrate the robust scalability of our framework. With Qwen3-1.7B, \ours achieves state-of-the-art results among previous methods with comparable model scale.

\paragraph{Comparison with previous token pruning methods.}
We make a comprehensive comparison with previous efficient methods of two representative categories under fair settings, \textit{i.e.}, implementation with Qwen3-0.6B. 
We adopt FLOPs to quantify the computational overhead, which is measured by a single forward pass with 2880 visual tokens and 100 text tokens.
As shown in \cref{tab:exp2}, previous methods sacrifice performance for the large compression ratio, where reducing tokens to 1/8 leads to significant performance drops. In contrast, \ours breaks through the dilemma by jointly considering the information \textit{capacity} and \textit{density}. Although with the increase of FLOPs, \ours significantly outperforms the previous methods while retaining only $7.5\%$ of visual tokens.

\subsection{Ablation Study}
In this section, we conduct analysis experiments on the widely used benchmarks using Qwen3-0.6B as the LLM backbone. The LLaVA-1.5 dataset~\cite{llava} is for model training, and the other training hyperparameters remain consistent by default. More results are deferred to the appendix.

\begin{table*}[t]
\begin{minipage}[c]{\textwidth}
    \hspace{0.025\textwidth}
    \begin{minipage}{0.48\textwidth}
    \makeatletter\def\@captype{table}
    \caption{Effectiveness of the main modules.}
    \vspace{-10pt}
    \centering
    \footnotesize
    \setlength{\tabcolsep}{2.6pt}
    \begin{tabular}{c c c | c | c c c c c}
    \toprule
    MKE & HTE & SIP & Token & MME & MMB & AI2D & TextVQA & Avg. \\
    \midrule
      &  & & 100\%  & 1318& 53.4& 48.8& 44.0& 48.3\\
    \checkmark &  &  & 32\% & 1400& 55.4& 49.2& 42.1& 49.1\\
    \checkmark & \checkmark  &  & 7.5\% & 1413& 54.9& 50.3& 
41.1& 
49.2\\
    \checkmark & \checkmark & \checkmark & \textbf{7.5\%} & 
1398 & 55.8 & 50.3 & 42.1 & \textbf{49.5}\\
    \bottomrule
    \end{tabular}
    
    \label{tab:ablation_modules}
    \end{minipage}
    \hspace{0.025\textwidth}
    \begin{minipage}{0.48\textwidth} %
    \makeatletter\def\@captype{table}
    \caption{Impacts of designs of HTE.}
    \vspace{-10pt}
    \centering
    \footnotesize
    \setlength{\tabcolsep}{2.6pt}
    \begin{tabular}{c|ccccc}
    \toprule
    HTE &MME & MMB & AI2D & TextVQA & Avg. \\
    \midrule
     First-Half &1276& 54.6& 48.6& 34.6& 45.8 \\
     Sparse &1383& 54.6& 49.5& 37.6& 47.8 \\
     Dense &1291& 54.3& 48.5& 36.6& 46.4 \\
     Second-Half &1398& 55.8& 50.3& 42.1& \textbf{49.5}\\
    \bottomrule
    \end{tabular}
    \label{tab:ablation_HTE}
    \end{minipage}
    \vfill
    \vspace{5pt}
    \hspace{0.025\textwidth}
    \begin{minipage}{0.5\textwidth}
    \makeatletter\def\@captype{table}
    \caption{Effect of different strategies of merging.}
    \vspace{-10pt}
    \centering
    \footnotesize
    \setlength{\tabcolsep}{2.8pt}
    \begin{tabular}{c|c|ccccc}
    \toprule
    Method & Token & MME & MMB & AI2D & TextVQA & Avg.\\
    \midrule
    MKE w/o. merge & 200\%& 1440& 56.9& 51.0 & 44.4& 50.9\\
    MKE w. self-merge & 25\%& 1317& 55.8& 48.7& 36.5& 47.0\\
    MKE & 32\%& 1398& 55.8& 50.3& 42.1& 49.5\\ 
    \bottomrule
    \end{tabular}
    \label{tab:ablation_tome}
    \end{minipage}
    \hspace{0.05\textwidth}
    \begin{minipage}{0.42\textwidth}
    \makeatletter\def\@captype{table}
    \caption{Impacts of different architectures of SIP.}
    \vspace{-10pt}
    \centering
    \footnotesize
    \setlength{\tabcolsep}{2.2pt}
    \begin{tabular}{c|ccccc}
    \toprule
    Method & MME & MMB & AI2D & TextVQA & Avg. \\
    \midrule
    w/o. SIP & 1413& 54.9& 50.3& 41.1& 49.2\\
    SIP + $\mathcal{L}_{rec}$ & 1328& 54.5& 48.4& 40.3
& 47.7\\ 
    SIP + $\mathcal{L}_{qrec}$ & 1398& 55.8& 50.3& 42.1& \textbf{49.5}\\ 
    \bottomrule
    \end{tabular}
    \label{tab:ablation_SAIP}
    \end{minipage}
\end{minipage}
\end{table*}
\begin{table*}[ht]
\centering
\footnotesize
\renewcommand\arraystretch{0.75}
\vspace{-5pt}
\setlength{\tabcolsep}{7.5pt}
\caption{The compatibility of HTE on MoE frameworks.}
\vspace{-10pt}
\begin{tabular}{cc|cc|cccccc}
\toprule
Backbone & HTE & Token & FLOPs(G) & MME & MMB &AI2D &TextVQA &GQA &Avg. \\
\midrule
Kimi-VL-16B-A3B & & 100\% & 7900 & 2195 & 80.8 & 82.6 & 91.9 & 60.0 & 78.7 \\
Kimi-VL-16B-A3B & \checkmark & 23\% & 6019 & 2208 & 80.7 & 82.6 & 91.1 & 59.8 & 78.6 \\
\midrule
VEN-VL-0.6B & & 32\% & 716 & 1424 & 56.4 & 50.3 & 42.5 & 54.8 & 50.9 \\
VEN-VL-0.6B & \checkmark & 7.5\% & 558 & 1413 & 54.9 & 50.3 & 41.1 & 54.4 & 50.2 \\
\bottomrule
\end{tabular}
\vspace{-15pt}
\label{tab:ablation_hte_compatibility2}
\end{table*}

\vspace{-5pt}
\paragraph{Component Analysis.}

We investigate the effectiveness of each component in \cref{tab:ablation_modules}. The integration of MKE brings the improvement of +0.8$\%$ on Avg. with the compression ratio of $68\%$. 
It validates the rationality of our design where enriched semantic modeling with heterogeneous priors and sequential merging for redundancy alleviation benefits the information capacity and diversity. 
HTE further progressively refines the visual distribution by pruning the token sequence to merely $7.5\%$ but with superior information density, \textit{e.g.},  $+1.1\%$ on AI2D, enabling effective and efficient visual perception. 
It is worth noting that the ablation of SIP leads to challenge in maintenance the valuable structure information, \textit{e.g.}, $-1.0\%$ in TextVQA.

\vspace{-5pt}
\paragraph{Variant of MKE design.}
We quantify the impact of different merging strategies within MKE in \cref{tab:ablation_tome}, It can be seen that although directly concatenating the multi-aspect features without merging yields the highest performance, the quartic complexity brought by $200\%$ is unacceptable. 
If operating both features with the individual self-merging strategy, it leads to drastic performance degradation despite the largest compression ratio. 
In contrast, our proposed MKE strategy strikes the optimal balance, \textit{e.g.}, 49.5\% on Avg. with 32\% tokens. This verifies the effectiveness of the structure-aware compression, which preserves the vanilla spatial information and formulates the unified visual space for complete semantic modeling.

\vspace{-5pt}
\paragraph{Analysis on the HTE.}
We make a comprehensive analysis of HTE in terms of its variants and compatibility in \cref{tab:ablation_HTE} and \cref{tab:ablation_hte_compatibility2}. 1) We notice that applying HTE in the shallow layers (First-Half) results in the lowest performance, particularly in fine-grained tasks like TextVQA. Similarly, applying HTE across all layers (Dense) or at sparse intervals yields suboptimal results, as the fine-grained semantics necessitate later layers for full comprehension. This motivates us to employ HTE on the deep layers (Second-Half), which indicates that with well-aligned semantic space, the potential of compacting the visual tokens can be unleashed. 2) We further verify the compatibility of our design with the current MoE model. As shown in \cref{tab:ablation_hte_compatibility2}, directly applying HTE on Kimi-VL~\cite{kimivl} even without training, the performance gap is narrow while the inference speed can be largely boosted. It well depicts the considerable potential for scaling to larger models for more efficiency.

\vspace{-5pt}
\paragraph{Analysis on the SIP.}
To verify the rationality of SIP, we showcase the impact of different supervision strategies involved in \cref{tab:ablation_SAIP}. 
As illustrated, the incompatibility incurs when directly regressing the exact value of original features via Mean Squared Error, \textit{i.e.}, SIP + $\mathcal{L}_{rec}$ as the reconstruction objective. 
It leads to a notable performance degradation, especially on MME \textit{i.e.}, dropped significantly to 1328. 
We consider that the tight per-value alignment constrains the model towards concentration on the trivial pattern rather than structural information, thus impeding the dynamic token pruning process. 
In contrast, we incorporate the channel quantization method ($\mathcal{L}_{qrec}$) to map the exact value into a low-level representation, which transforms the supervision process from the per-token alignment into a more topology-based guidance.
Therefore, benefiting from the semantically alignment with the vanilla structure, \ours yields the most robust performance (Avg. 49.5) with the precise perception and improved reasoning ability.

\section{Conclusion}
This paper proposes \ours, a visual ensemble MoE framework aiming to bridge the gap between performance and efficiency for multi-modal understanding. Addressing the critical limitations of information capacity and density, \ours follows the \textit{enrich then compact} principle to information-enriched visual space via Multi-Aspect Knowledge Ensemble (MKE). Subsequently, through the Hierarchical Token Ensemble (HTE), we leverage adaptive MoE routers to progressively refine the visual distribution, enabling dynamic and precise token selection within the LLM layers for compacting. To guarantee semantic integrity during this reduction process, we empower the framework with Structure Information Preservation (SIP) via explicit visual supervision. Experimental results demonstrate that \ours achieves superior performance on complex multimodal tasks with significantly fewer but information-condensed tokens and great compatibility to scaling to larger models.

\section*{Limitations}

Despite the algorithmic optimization of VEN-VL, our empirical evaluations are currently constrained to small-scale models due to limited computational resources, and scaling this framework to massive state-of-the-art LLMs remains for future work. Additionally, while our approach significantly reduces theoretical FLOPs and visual token counts, translating these gains into actual end-to-end acceleration in real-world deployment further relies on system-level efforts, such as parallel deployment strategies and customized kernel optimizations.

\bibliography{custom}

@article{llava,
  title={Visual instruction tuning},
  author={Liu, Haotian and Li, Chunyuan and Wu, Qingyang and Lee, Yong Jae},
  journal={NeurIPS},
  volume={36},
  pages={34892--34916},
  year={2023}
}

@inproceedings{atpllava,
  title={Atp-llava: Adaptive token pruning for large vision language models},
  author={Ye, Xubing and Gan, Yukang and Ge, Yixiao and Zhang, Xiao-Ping and Tang, Yansong},
  booktitle={CVPR},
  pages={24972--24982},
  year={2025}
}

@article{instructblip,
  title={Instructblip: Towards general-purpose vision-language models with instruction tuning},
  author={Dai, Wenliang and Li, Junnan and Li, Dongxu and Tiong, Anthony and Zhao, Junqi and Wang, Weisheng and Li, Boyang and Fung, Pascale N and Hoi, Steven},
  journal={NeurIPS},
  volume={36},
  pages={49250--49267},
  year={2023}
}

@inproceedings{du2022glm,
  title={Glm: General language model pretraining with autoregressive blank infilling},
  author={Du, Zhengxiao and Qian, Yujie and Liu, Xiao and Ding, Ming and Qiu, Jiezhong and Yang, Zhilin and Tang, Jie},
  booktitle={ACL},
  pages={320--335},
  year={2022}
}

@inproceedings{blip2,
  title={Blip-2: Bootstrapping language-image pre-training with frozen image encoders and large language models},
  author={Li, Junnan and Li, Dongxu and Savarese, Silvio and Hoi, Steven},
  booktitle={ICML},
  pages={19730--19742},
  year={2023},
  organization={PMLR}
}

@article{minigpt4,
  title={Minigpt-4: Enhancing vision-language understanding with advanced large language models},
  author={Zhu, Deyao and Chen, Jun and Shen, Xiaoqian and Li, Xiang and Elhoseiny, Mohamed},
  journal={arXiv preprint arXiv:2304.10592},
  year={2023}
}

@inproceedings{visionzip,
  title={Visionzip: Longer is better but not necessary in vision language models},
  author={Yang, Senqiao and Chen, Yukang and Tian, Zhuotao and Wang, Chengyao and Li, Jingyao and Yu, Bei and Jia, Jiaya},
  booktitle={CVPR},
  pages={19792--19802},
  year={2025}
}

@inproceedings{fastv,
  title={An image is worth 1/2 tokens after layer 2: Plug-and-play inference acceleration for large vision-language models},
  author={Chen, Liang and Zhao, Haozhe and Liu, Tianyu and Bai, Shuai and Lin, Junyang and Zhou, Chang and Chang, Baobao},
  booktitle={ECCV},
  pages={19--35},
  year={2024},
  organization={Springer}
}

@article{pyramiddrop,
  title={Pyramiddrop: Accelerating your large vision-language models via pyramid visual redundancy reduction},
  author={Xing, Long and Huang, Qidong and Dong, Xiaoyi and Lu, Jiajie and Zhang, Pan and Zang, Yuhang and Cao, Yuhang and He, Conghui and Wang, Jiaqi and Wu, Feng and others},
  journal={arXiv preprint arXiv:2410.17247},
  year={2024}
}

@article{llava-mini,
  title={Llava-mini: Efficient image and video large multimodal models with one vision token},
  author={Zhang, Shaolei and Fang, Qingkai and Yang, Zhe and Feng, Yang},
  journal={arXiv preprint arXiv:2501.03895},
  year={2025}
}

@inproceedings{voco,
  title={Voco-llama: Towards vision compression with large language models},
  author={Ye, Xubing and Gan, Yukang and Huang, Xiaoke and Ge, Yixiao and Tang, Yansong},
  booktitle={CVPR},
  pages={29836--29846},
  year={2025}
}

@article{shazeer2017outrageously,
  title={Outrageously large neural networks: The sparsely-gated mixture-of-experts layer},
  author={Shazeer, Noam and Mirhoseini, Azalia and Maziarz, Krzysztof and Davis, Andy and Le, Quoc and Hinton, Geoffrey and Dean, Jeff},
  journal={ICLR},
  year={2017}
}

@article{eigen2013learning,
  title={Learning factored representations in a deep mixture of experts},
  author={Eigen, David and Ranzato, Marc'Aurelio and Sutskever, Ilya},
  journal={arXiv preprint arXiv:1312.4314},
  year={2013}
}

@article{llava-moe,
  title={Moe-llava: Mixture of experts for large vision-language models},
  author={Lin, Bin and Tang, Zhenyu and Ye, Yang and Cui, Jiaxi and Zhu, Bin and Jin, Peng and Huang, Jinfa and Zhang, Junwu and Pang, Yatian and Ning, Munan and others},
  journal={arXiv preprint arXiv:2401.15947},
  year={2024}
}

@article{kimivl,
  title={Kimi-vl technical report},
  author={Team, Kimi and Du, Angang and Yin, Bohong and Xing, Bowei and Qu, Bowen and Wang, Bowen and Chen, Cheng and Zhang, Chenlin and Du, Chenzhuang and Wei, Chu and others},
  journal={arXiv preprint arXiv:2504.07491},
  year={2025}
}

@inproceedings{molmo,
  title={Molmo and pixmo: Open weights and open data for state-of-the-art vision-language models},
  author={Deitke, Matt and Clark, Christopher and Lee, Sangho and Tripathi, Rohun and Yang, Yue and Park, Jae Sung and Salehi, Mohammadreza and Muennighoff, Niklas and Lo, Kyle and Soldaini, Luca and others},
  booktitle={CVPR},
  pages={91--104},
  year={2025}
}

@article{deepseek-vl2,
  title={Deepseek-vl2: Mixture-of-experts vision-language models for advanced multimodal understanding},
  author={Wu, Zhiyu and Chen, Xiaokang and Pan, Zizheng and Liu, Xingchao and Liu, Wen and Dai, Damai and Gao, Huazuo and Ma, Yiyang and Wu, Chengyue and Wang, Bingxuan and others},
  journal={arXiv preprint arXiv:2412.10302},
  year={2024}
}

@inproceedings{clip,
  title={Learning transferable visual models from natural language supervision},
  author={Radford, Alec and Kim, Jong Wook and Hallacy, Chris and Ramesh, Aditya and Goh, Gabriel and Agarwal, Sandhini and Sastry, Girish and Askell, Amanda and Mishkin, Pamela and Clark, Jack and others},
  booktitle={ICML},
  pages={8748--8763},
  year={2021},
  organization={PmLR}
}

@article{tome,
  title={Token merging: Your vit but faster},
  author={Bolya, Daniel and Fu, Cheng-Yang and Dai, Xiaoliang and Zhang, Peizhao and Feichtenhofer, Christoph and Hoffman, Judy},
  journal={arXiv preprint arXiv:2210.09461},
  year={2022}
}

@inproceedings{heinrich2025radiov2,
  title={Radiov2. 5: Improved baselines for agglomerative vision foundation models},
  author={Heinrich, Greg and Ranzinger, Mike and Yin, Hongxu and Lu, Yao and Kautz, Jan and Tao, Andrew and Catanzaro, Bryan and Molchanov, Pavlo},
  booktitle={CVPR},
  pages={22487--22497},
  year={2025}
}

@article{llama3,
  title={The llama 3 herd of models},
  author={Grattafiori, Aaron and Dubey, Abhimanyu and Jauhri, Abhinav and Pandey, Abhinav and Kadian, Abhishek and Al-Dahle, Ahmad and Letman, Aiesha and Mathur, Akhil and Schelten, Alan and Vaughan, Alex and others},
  journal={arXiv preprint arXiv:2407.21783},
  year={2024}
}

@article{openai2023gpt4,
  title={{GPT}-4 technical report},
  author={OpenAI},
  journal={arXiv preprint arXiv:2303.08774},
  year={2023}
}

@article{qwen3,
  title={Qwen3 technical report},
  author={Yang, An and Li, Anfeng and Yang, Baosong and Zhang, Beichen and Hui, Binyuan and Zheng, Bo and Yu, Bowen and Gao, Chang and Huang, Chengen and Lv, Chenxu and others},
  journal={arXiv preprint arXiv:2505.09388},
  year={2025}
}

@article{qwen2vl,
  title={Qwen2-vl: Enhancing vision-language model's perception of the world at any resolution},
  author={Wang, Peng and Bai, Shuai and Tan, Sinan and Wang, Shijie and Fan, Zhihao and Bai, Jinze and Chen, Keqin and Liu, Xuejing and Wang, Jialin and Ge, Wenbin and others},
  journal={arXiv preprint arXiv:2409.12191},
  year={2024}
}

@article{gemini25,
  title={Gemini 2.5: Pushing the frontier with advanced reasoning, multimodality, long context, and next generation agentic capabilities},
  author={Comanici, Gheorghe and Bieber, Eric and Schaekermann, Mike and Pasupat, Ice and Sachdeva, Noveen and Dhillon, Inderjit and Blistein, Marcel and Ram, Ori and Zhang, Dan and Rosen, Evan and others},
  journal={arXiv preprint arXiv:2507.06261},
  year={2025}
}

@article{sparsevlm,
  title={Sparsevlm: Visual token sparsification for efficient vision-language model inference},
  author={Zhang, Yuan and Fan, Chun-Kai and Ma, Junpeng and Zheng, Wenzhao and Huang, Tao and Cheng, Kuan and Gudovskiy, Denis and Okuno, Tomoyuki and Nakata, Yohei and Keutzer, Kurt and others},
  journal={arXiv preprint arXiv:2410.04417},
  year={2024}
}

@inproceedings{mmbench,
  title={Mmbench: Is your multi-modal model an all-around player?},
  author={Liu, Yuan and Duan, Haodong and Zhang, Yuanhan and Li, Bo and Zhang, Songyang and Zhao, Wangbo and Yuan, Yike and Wang, Jiaqi and He, Conghui and Liu, Ziwei and others},
  booktitle={ECCV},
  pages={216--233},
  year={2024},
}

@article{mme,
  author       = {Chaoyou Fu and
                  Peixian Chen and
                  Yunhang Shen and
                  Yulei Qin and
                  Mengdan Zhang and
                  Xu Lin and
                  Zhenyu Qiu and
                  Wei Lin and
                  Jinrui Yang and
                  Xiawu Zheng and
                  Ke Li and
                  Xing Sun and
                  Rongrong Ji},
  title        = {{MME:} {A} Comprehensive Evaluation Benchmark for Multimodal Large
                  Language Models},
  journal={arXiv preprint arXiv:2306.13394},
  year={2024},
}

@inproceedings{textvqa,
  title={Towards vqa models that can read},
  author={Singh, Amanpreet and Natarajan, Vivek and Shah, Meet and Jiang, Yu and Chen, Xinlei and Batra, Dhruv and Parikh, Devi and Rohrbach, Marcus},
  booktitle={CVPR},
  pages={8317--8326},
  year={2019}
}

@inproceedings{gqa,
  title={Gqa: A new dataset for real-world visual reasoning and compositional question answering},
  author={Hudson, Drew A and Manning, Christopher D},
  booktitle={CVPR},
  pages={6700--6709},
  year={2019}
}

@inproceedings{sharegpt4v,
  title={Sharegpt4v: Improving large multi-modal models with better captions},
  author={Chen, Lin and Li, Jinsong and Dong, Xiaoyi and Zhang, Pan and He, Conghui and Wang, Jiaqi and Zhao, Feng and Lin, Dahua},
  booktitle={ECCV},
  pages={370--387},
  year={2024},
  organization={Springer}
}

@inproceedings{docvqa,
  title={Docvqa: A dataset for vqa on document images},
  author={Mathew, Minesh and Karatzas, Dimosthenis and Jawahar, CV},
  booktitle={WACV},
  pages={2200--2209},
  year={2021}
}

@inproceedings{ai2d,
  title={A diagram is worth a dozen images},
  author={Kembhavi, Aniruddha and Salvato, Mike and Kolve, Eric and Seo, Minjoon and Hajishirzi, Hannaneh and Farhadi, Ali},
  booktitle={ECCV},
  pages={235--251},
  year={2016},
  organization={Springer}
}

@article{mathvista,
  title={Mathvista: Evaluating mathematical reasoning of foundation models in visual contexts},
  author={Lu, Pan and Bansal, Hritik and Xia, Tony and Liu, Jiacheng and Li, Chunyuan and Hajishirzi, Hannaneh and Cheng, Hao and Chang, Kai-Wei and Galley, Michel and Gao, Jianfeng},
  journal={arXiv preprint arXiv:2310.02255},
  year={2023}
}

@inproceedings{mmmu,
  title={Mmmu: A massive multi-discipline multimodal understanding and reasoning benchmark for expert agi},
  author={Yue, Xiang and Ni, Yuansheng and Zhang, Kai and Zheng, Tianyu and Liu, Ruoqi and Zhang, Ge and Stevens, Samuel and Jiang, Dongfu and Ren, Weiming and Sun, Yuxuan and others},
  booktitle={CVPR},
  pages={9556--9567},
  year={2024}
}

@inproceedings{uvcom,
  title={Bridging the gap: A unified video comprehension framework for moment retrieval and highlight detection},
  author={Xiao, Yicheng and Luo, Zhuoyan and Liu, Yong and Ma, Yue and Bian, Hengwei and Ji, Yatai and Yang, Yujiu and Li, Xiu},
  booktitle={CVPR},
  pages={18709--18719},
  year={2024}
}

@inproceedings{weaklysupervised,
  title={Weakly supervised temporal action localization via representative snippet knowledge propagation},
  author={Huang, Linjiang and Wang, Liang and Li, Hongsheng},
  booktitle={CVPR},
  pages={3272--3281},
  year={2022}
}

@article{random_walk,
  title={Random walk graph neural networks},
  author={Nikolentzos, Giannis and Vazirgiannis, Michalis},
  journal={NeurIPS},
  volume={33},
  pages={16211--16222},
  year={2020}
}

@article{stablerank,
  title={Sampling from large matrices: An approach through geometric functional analysis},
  author={Rudelson, Mark and Vershynin, Roman},
  journal={Journal of the ACM (JACM)},
  volume={54},
  number={4},
  pages={21--es},
  year={2007},
  publisher={ACM New York, NY, USA}
}

@article{coderate,
  title={Learning diverse and discriminative representations via the principle of maximal coding rate reduction},
  author={Yu, Yaodong and Chan, Kwan Ho Ryan and You, Chong and Song, Chaobing and Ma, Yi},
  journal={NeurIPS},
  volume={33},
  pages={9422--9434},
  year={2020}
}

@misc{liu2024llavanext,
    title={LLaVA-NeXT: Improved reasoning, OCR, and world knowledge},
    url={https://llava-vl.github.io/blog/2024-01-30-llava-next/},
    author={Liu, Haotian and Li, Chunyuan and Li, Yuheng and Li, Bo and Zhang, Yuanhan and Shen, Sheng and Lee, Yong Jae},
    month={January},
    year={2024}
}

@article{team2024qwen2,
  title={Qwen2 technical report},
  author={Team, Qwen and others},
  journal={arXiv preprint arXiv:2407.10671},
  volume={2},
  number={3},
  year={2024}
}

@article{llavamod,
  title={Llava-mod: Making llava tiny via moe knowledge distillation},
  author={Shu, Fangxun and Liao, Yue and Zhuo, Le and Xu, Chenning and Zhang, Lei and Zhang, Guanghao and Shi, Haonan and Chen, Long and Zhong, Tao and He, Wanggui and others},
  journal={arXiv preprint arXiv:2408.15881},
  year={2024}
}

@article{ross,
  title={Reconstructive visual instruction tuning},
  author={Wang, Haochen and Zheng, Anlin and Zhao, Yucheng and Wang, Tiancai and Ge, Zheng and Zhang, Xiangyu and Zhang, Zhaoxiang},
  journal={arXiv preprint arXiv:2410.09575},
  year={2024}
}

@article{fsq,
  title={Finite scalar quantization: Vq-vae made simple},
  author={Mentzer, Fabian and Minnen, David and Agustsson, Eirikur and Tschannen, Michael},
  journal={arXiv preprint arXiv:2309.15505},
  year={2023}
}

@inproceedings{prumerge,
  title={Llava-prumerge: Adaptive token reduction for efficient large multimodal models},
  author={Shang, Yuzhang and Cai, Mu and Xu, Bingxin and Lee, Yong Jae and Yan, Yan},
  booktitle={CVPR},
  pages={22857--22867},
  year={2025}
}

@article{seedbench,
  title={Seed-bench: Benchmarking multimodal llms with generative comprehension},
  author={Li, Bohao and Wang, Rui and Wang, Guangzhi and Ge, Yuying and Ge, Yixiao and Shan, Ying},
  journal={arXiv preprint arXiv:2307.16125},
  year={2023}
}

@inproceedings{avit,
  title={A-vit: Adaptive tokens for efficient vision transformer},
  author={Yin, Hongxu and Vahdat, Arash and Alvarez, Jose M and Mallya, Arun and Kautz, Jan and Molchanov, Pavlo},
  booktitle={CVPR},
  pages={10809--10818},
  year={2022}
}

@inproceedings{spvit,
  title={Spvit: Enabling faster vision transformers via latency-aware soft token pruning},
  author={Kong, Zhenglun and Dong, Peiyan and Ma, Xiaolong and Meng, Xin and Niu, Wei and Sun, Mengshu and Shen, Xuan and Yuan, Geng and Ren, Bin and Tang, Hao and others},
  booktitle={ECCV},
  pages={620--640},
  year={2022},
  organization={Springer}
}

@inproceedings{clevrmath,
  title={Clevr: A diagnostic dataset for compositional language and elementary visual reasoning},
  author={Johnson, Justin and Hariharan, Bharath and Van Der Maaten, Laurens and Fei-Fei, Li and Lawrence Zitnick, C and Girshick, Ross},
  booktitle={Proceedings of the IEEE conference on computer vision and pattern recognition},
  pages={2901--2910},
  year={2017}
}

@inproceedings{dvqa,
  title={Dvqa: Understanding data visualizations via question answering},
  author={Kafle, Kushal and Price, Brian and Cohen, Scott and Kanan, Christopher},
  booktitle={Proceedings of the IEEE conference on computer vision and pattern recognition},
  pages={5648--5656},
  year={2018}
}

@inproceedings{okvqa,
  title={Ok-vqa: A visual question answering benchmark requiring external knowledge},
  author={Marino, Kenneth and Rastegari, Mohammad and Farhadi, Ali and Mottaghi, Roozbeh},
  booktitle={Proceedings of the IEEE/cvf conference on computer vision and pattern recognition},
  pages={3195--3204},
  year={2019}
}

@article{docmatix,
  title={Building and better understanding vision-language models: insights and future directions},
  author={Lauren{\c{c}}on, Hugo and Marafioti, Andr{\'e}s and Sanh, Victor and Tronchon, L{\'e}o},
  journal={arXiv preprint arXiv:2408.12637},
  year={2024}
}

@article{sciqa,
  title={Learn to explain: Multimodal reasoning via thought chains for science question answering},
  author={Lu, Pan and Mishra, Swaroop and Xia, Tanglin and Qiu, Liang and Chang, Kai-Wei and Zhu, Song-Chun and Tafjord, Oyvind and Clark, Peter and Kalyan, Ashwin},
  journal={Advances in Neural Information Processing Systems},
  volume={35},
  pages={2507--2521},
  year={2022}
}

@inproceedings{pope,
  title={Evaluating Object Hallucination in Large Vision-Language Models},
  author={Li, Yifan and Du, Yifan and Zhou, Kun and Wang, Jinpeng and Zhao, Wayne Xin and Wen, Ji-Rong},
  booktitle={Proceedings of the 2023 Conference on Empirical Methods in Natural Language Processing},
  pages={292--305},
  year={2023}
}

@article{star,
  title={STAR: Stage-Wise Attention-Guided Token Reduction for Efficient Large Vision-Language Models Inference},
  author={Guo, Yichen and Li, Hanze and Zhang, Zonghao and You, Jinhao and Tang, Kai and Huang, Xiande},
  journal={arXiv preprint arXiv:2505.12359},
  year={2025}
}

\newpage

\appendix

\section{Appendix}
\label{sec:appendix}

\subsection{Experiment Setup}
\label{subsec:exp_setup}

\paragraph{Implementation Details.} 
We follow the similar architecture of LLaVA series~\cite{liu2024llavanext, llava}. In terms of the visual extractor, CLIP-ViT-L/14-336px\cite{clip} and RADIO-v2.5\cite{heinrich2025radiov2} are leveraged to obtain multiple-aspect visual features for semantic knowledge ensemble. To effectively process variable-resolution inputs, we employ the adaptive image tiling strategy. For the language backbone, we conduct experiments on Qwen2/3 family~\cite{team2024qwen2,qwen3} of different parameter sizes with MoE design to verify our scalability. 

\paragraph{Training Strategy.} 
We follow a standard two-stage training pipeline. In the initial Vision-Language Alignment stage, we freeze the vision encoders and LLM, training only the projection layers with a learning rate of $2e^{-4}$ and a batch size of 1024. In the subsequent SFT stage, we make the entire model trainable with a learning rate of $4e^{-5}$. Notably, unlike prior MoE methods~\cite{llava-moe,llavamod} that require a separate warm-up stage for experts, we integrate the MoE optimization directly into the SFT phase for simplicity. The overall training objective integrates the next-token-prediction loss $\mathcal{L}_{ntp}$ with the SIP guidance $\mathcal{L}_{qrec}$ and the MoE auxiliary load balancing loss. By default, the activated visual experts are configured with an intermediate dimension of half the original FFN size to maintain parameter efficiency.

\paragraph{Datasets \& Benchmarks.}
We train our model based on publicly multi-modal understanding datasets. The alignment stage leverages substantial image-text pairs, \textit{e.g.}, LCS-558K~\cite{llava} for concept learning. The SFT stage employs a diverse mix of 10M instruction-tuning data, spanning across general visual understanding, \textit{e.g.}, ShareGPT4V~\cite{sharegpt4v}, OKVQA~\cite{okvqa}, OCR and documents, \textit{e.g.}, DocVQA~\cite{docvqa}, Docmatix ~\cite{docmatix}, chart$\&$diagram QA, \textit{e.g.}, AI2D~\cite{ai2d}, DVQA~\cite{dvqa} and reasoning tasks, \textit{e.g.}, MathVista~\cite{mathvista}, CLEVER-MATH~\cite{clevrmath}, augmented with pure-text instruction data to preserve language capabilities. 
For evaluation, we conduct extensive evaluation across a wide range of benchmarks, including General Understanding (MMBench~\cite{mmbench}, MME~\cite{mme}, SEEDBench~\cite{seedbench}), Fine-grained Perception (TextVQA~\cite{textvqa}, GQA~\cite{gqa}), Hallucination (POPE~\cite{pope}), and Knowledge and Reasoning (SciQA~\cite{sciqa}, AI2D~\cite{ai2d}, MMMU~\cite{mmmu}).

\subsection{Additional Analyses and Ablations}

\paragraph{Analysis on HTE's token selection method.}

In \cref{tab:ablation_tokenselection}, we describe the reduction-gated mechanism in HTE by comparing different token scoring functions with \cref{eq:token_score}. Random selection strategy serves as a baseline, resulting in significant performance deterioration, \textit{e.g.}, -107 on MME and -2.0\% on MMB, which underscores that visual tokens in deep layers are not uniformly redundant and indicates that the content-aware pruning is essential. 

Strategies utilizing routing score consistently outperform the random baseline, with Sum achieving a competitive average of 49.3\% and Max yielding the optimal performance of 49.5\%. It validates our core premise that routing scores effectively serve as indicators of token importance: high-value tokens elicit strong, confident activation of specific experts, whereas redundant tokens exhibit flatter distributions. 
Furthermore, the superiority of Max suggests that identifying the peak expert demand rather than the accumulated total is the most precise metric for preserving valuable tokens that require specialized processing.

\begin{table}[h]
\centering
\footnotesize
\vspace{-5pt}
\setlength{\tabcolsep}{4.5pt}
\caption{Comparison of token selection methods.}
\vspace{-5pt}
\begin{tabular}{c|ccccc}
\toprule
Token Score & MME & MMB & AI2D & TextVQA & Avg.\\ 
\midrule
Random & 1291& 53.8& 47.8& 40.6& 47.1\\
Sum  & 1395& 55.4& 49.9& 42.2& 49.3\\ 
Max & 1398& 55.8& 50.3& 42.1& \textbf{49.5}\\
\bottomrule
\end{tabular}
\vspace{-10pt}
\label{tab:ablation_tokenselection}
\end{table}

\begin{figure}[h]
\centering
\includegraphics[width=\linewidth]{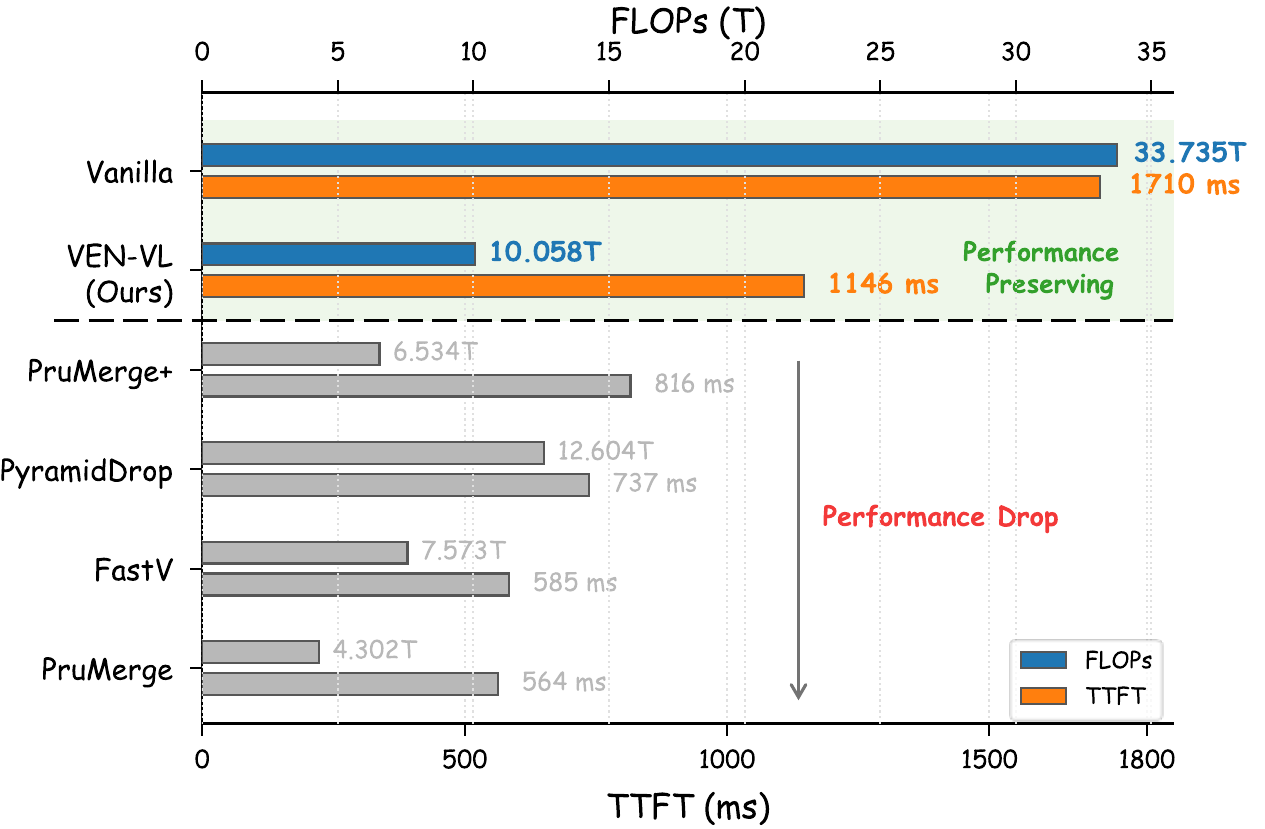}
\caption{Real-world inference cost on Qwen3-8B. TTFT denotes time-to-first-token.}
\label{fig:latency_qwen8b}
\end{figure}

\paragraph{Real-world Inference Latency.}
We further evaluate the end-to-end inference latency on Qwen3-8B to provide a direct hardware measurement beyond FLOPs. Specifically, we report the time-to-first-token (TTFT) under the same evaluation setting in \cref{fig:latency_qwen8b}. Compared with the Vanilla baseline, \ours reduces the visual tokens to $7.5\%$ and decreases TTFT from \textbf{1710ms} to \textbf{1146ms}, indicating that the compact visual representation can bring practical acceleration over the standard dense visual sequence. Meanwhile, due to the additional visual branch introduced by MKE, \ours still has higher wall-clock latency than particular pruning methods such as PruMerge or FastV. However, we argue that simply sacrificing performance for acceleration is unsatisfactory. Although our design introduces additional computational overhead compared with these methods, it yields clear performance improvements. Furthermore, owing to its strong scalability to larger MoE models, as validated in \cref{tab:ablation_hte_compatibility2}, our method offers a more promising solution for real-world deployment for effective and efficient multi-modal understanding.

\begin{figure}[h]
\centering
\includegraphics[width=\linewidth]{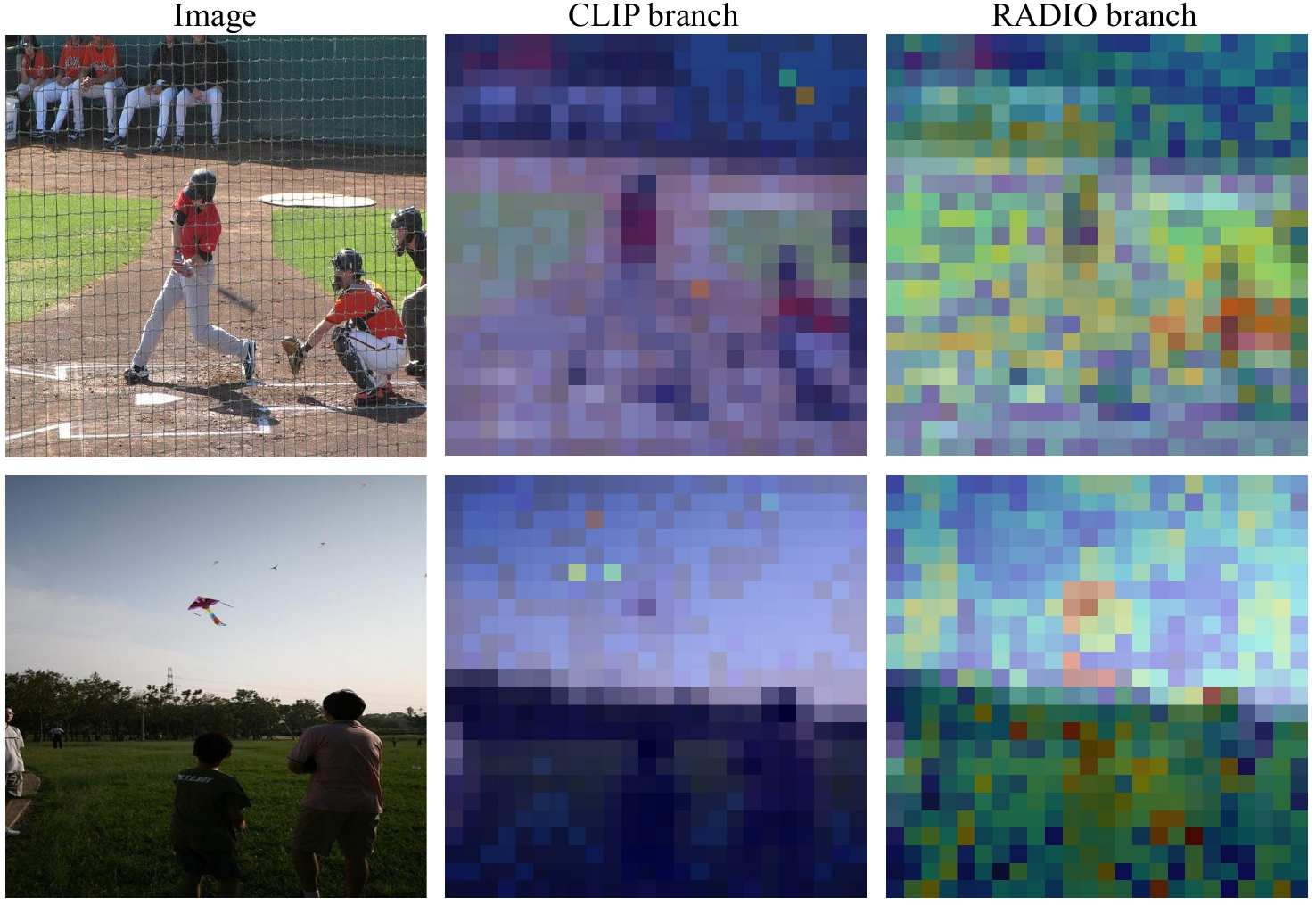}
\caption{Visualization of multi-aspect visual features from CLIP and RADIO branches.}
\label{fig:mke_visualization}
\end{figure}

\paragraph{Visualization of Multi-aspect Visual Features.}
We visualize the feature responses from the two visual branches to further understand the complementarity of MKE. As shown in \cref{fig:mke_visualization}, the CLIP branch tends to focus on salient objects and semantic regions, while the RADIO branch preserves more fine-grained textures and local patterns. This observation supports our motivation that multi-aspect visual features provide complementary visual clues, rather than repeated encoding from two homogeneous views. By unifying these heterogeneous representations with MKE, \ours obtains a visual representation with improved information capacity before the subsequent token refinement.

\paragraph{Individual effect of HTE and MKE.}
We conduct additional ablation to verify each of the effect of HTE and MKE in \cref{tab:disentangle_hte_mke}. With the single-encoder visual input, HTE only improves the average score from 48.3 to 50.9, which indicates that the MoE router itself can serve as an effective indicator to identify information-condensed tokens and improve information density. However, the FLOPs remain high because the input visual sequence still contains substantial redundancy. After incorporating MKE, the multi-aspect visual features are compressed into a more enriched and compact representation, reducing FLOPs from 1872G to 558G while maintaining comparable performance. Finally, the full \ours further introduces SIP to preserve structure information, bringing the best average performance under the same token ratio and computational cost.

\begin{table}[h]
\centering
\footnotesize
\setlength{\tabcolsep}{3.0pt}
\caption{Individual effect of HTE and MKE.}
\label{tab:disentangle_hte_mke}
\resizebox{\columnwidth}{!}{%
\begin{tabular}{c|cc|ccccc}
\toprule
Method & Tokens & FLOPs(G) & MME & MMB & AI2D & TextVQA & Avg. \\
\midrule
Vanilla & 100\% & 2330 & 1318 & 53.4 & 48.8 & 44.0 & 48.3 \\
HTE only & 22.8\% & 1872 & 1409 & 56.7 & 50.9 & 45.8 & 50.9 \\
MKE+HTE & 7.5\% & 558 & 1413 & 54.9 & 50.3 & 41.1 & 49.2 \\
\ours & 7.5\% & 558 & 1398 & 55.8 & 50.3 & 42.1 & 49.5 \\
\bottomrule
\end{tabular}
}
\end{table}

\paragraph{Parameter Analysis of DropRate.}
We analyze the sensitivity of DropRate in HTE across three representative benchmarks, including MMBench for natural images, OCRBench for OCR documents, and AI2D for chart and diagram understanding. As shown in \cref{tab:droprate_sensitivity}, DropRate controls the effective compression strength of HTE. When DropRate is no larger than 0.1, the performance remains relatively stable while the number of visual tokens and FLOPs are significantly reduced. However, further increasing DropRate to 0.15 or 0.2 leads to performance degradation, especially on OCRBench and AI2D, where fine-grained visual details are critical. Therefore, we adopt DropRate $=0.1$ as the default setting to balance performance and efficiency.

\begin{table}[h]
\centering
\footnotesize
\setlength{\tabcolsep}{2.5pt}
\caption{Sensitivity analysis of DropRate in HTE.}
\label{tab:droprate_sensitivity}
\resizebox{\columnwidth}{!}{%
\begin{tabular}{c|ccccc}
\toprule
DropRate & Final Tokens & FLOPs(G) & MMBench & OCRBench & AI2D \\
\midrule
-- & 100\% & 2330 & 53.4 & 34.0 & 48.8 \\
0.0 & 32\% & 716 & 56.4 & 33.8 & 50.3 \\
0.05 & 16\% & 618 & 56.8 & 32.7 & 50.9 \\
0.1 & 7.5\% & 558 & 55.8 & 33.0 & 50.3 \\
0.15 & 3.3\% & 519 & 55.0 & 31.2 & 49.5 \\
0.2 & 1.4\% & 492 & 53.7 & 30.4 & 47.9 \\
\bottomrule
\end{tabular}
}
\end{table}

\paragraph{Parameter Analysis of MKE Merge Ratios.}
We further study the merge ratios in MKE, where $k_s$ denotes the self-merge ratio and $k_c$ denotes the cross-merge ratio. As shown in \cref{tab:mke_merge_sensitivity}, directly concatenating two visual branches improves the performance by enlarging information capacity, but it also doubles the input tokens and brings heavy computational overhead. In contrast, moderate merging ratios can effectively suppress redundancy while preserving multi-aspect information. When the merging becomes too aggressive, the performance on MMBench and OCRBench decreases, indicating that excessive merging may damage semantic regions and fine-grained text patterns. We set $k_s=0.5$ and $k_c=0.4$ by default, which provides a favorable balance between compactness and visual understanding.

\begin{table}[h]
\centering
\footnotesize
\setlength{\tabcolsep}{2.5pt}
\caption{Sensitivity analysis of MKE merge ratios.}
\label{tab:mke_merge_sensitivity}
\resizebox{\columnwidth}{!}{%
\begin{tabular}{c|cc|ccccc}
\toprule
Setting & $k_s$ & $k_c$ & Input Tokens & FLOPs(G) & MMBench & OCRBench & AI2D \\
\midrule
Vanilla & -- & -- & 100\% & 2330 & 53.4 & 34.0 & 48.8 \\
w/o. merge & -- & -- & 200\% & 5297 & 56.9 & 34.2 & 51.0 \\
MKE & 0.0 & 0.0 & 50\% & 884 & 56.3 & 33.3 & 49.6 \\
MKE & 0.3 & 0.2 & 39\% & 674 & 56.0 & 32.7 & 50.1 \\
MKE & 0.3 & 0.4 & 35\% & 611 & 55.9 & 33.0 & 49.9 \\
MKE & 0.5 & 0.4 & 32\% & 558 & 55.8 & 33.0 & 50.3 \\
MKE & 0.8 & 0.4 & 28\% & 480 & 53.7 & 31.5 & 49.2 \\
MKE & 0.8 & 0.6 & 27\% & 463 & 53.3 & 30.2 & 48.7 \\
\bottomrule
\end{tabular}
}
\end{table}

\paragraph{Parameter Analysis of Semantic Propagation Factor.}
We also evaluate the propagation factor $\omega$ in SIP. Different from DropRate and merge ratios, $\omega$ does not change the number of visual tokens or FLOPs, but controls the strength of structure information propagation. As shown in \cref{tab:omega_sensitivity}, $\omega=0.8$ achieves the best 6-Avg. score, while smaller or larger values slightly degrade the overall performance. Therefore, we use $\omega=0.8$ as the default setting in our experiments.

\begin{table}[h]
\centering
\footnotesize
\setlength{\tabcolsep}{3.0pt}
\caption{Sensitivity analysis of the propagation factor.}
\label{tab:omega_sensitivity}
\begin{tabular}{c|cccc}
\toprule
$\omega$ & MMBench & OCRBench & AI2D & 6-Avg. \\
\midrule
0.7 & 56.0 & 33.0 & 48.8 & 47.6 \\
0.8 & 55.8 & 33.0 & 50.3 & 47.9 \\
0.9 & 55.6 & 32.6 & 50.7 & 47.4 \\
\bottomrule
\end{tabular}
\end{table}

\end{document}